# Shifting social norms as a driving force for linguistic change: Struggles about language and gender in the German Bundestag[1]

**Authors:** Carolin Müller-Spitzer[a]*, Samira Ochs[a]

[a] *Leibniz Institute for the German Language (IDS), Mannheim, Germany*

*\* Corresponding author*

> „Die Frauenrechtler mögen verzweifeln, aber es läßt sich nun einmal nicht ändern: die Sprache hält's mit dem Mann. Sie ist noch immer nicht emanzipiert."[2] Karl Kraus (1912)

**Abstract**

This paper focuses on language change based on shifting social norms, in particular with regard to the debate on language and gender. It is a recurring argument in this debate that language develops 'naturally' and that 'severe interventions' - such as gender-inclusive language is often claimed to be – in the allegedly 'organic' language system are inappropriate and even 'dangerous'. Such 'interventions' are, however, not unprecedented. Socially motivated processes of language change are neither unusual nor new. We focus in our contribution on one important political-social space in Germany, the German Bundestag. Taking other struggles about language and gender in the plenaries of the Bundestag as a starting point, our article illustrates that language and gender has been a recurring issue in the German Bundestag since the 1980s. We demonstrate how this is reflected in linguistic practices of the Bundestag, by the use of a) designations for gays and lesbians; b) pair forms such as *Bürgerinnen und Bürger* ('female and male citizens'); and c) female forms of addresses and personal nouns (*Präsidentin* in addition to *Präsident*). Lastly, we will discuss implications of these earlier language battles for the currently very heated debate about gender-inclusive language, especially regarding new forms with gender symbols like the asterisk or the colon (*Lehrer\*innen, Lehrer:innen;* [male\*female teachers]) which are intended to encompass all gender identities.

---

[1] We thank our IDS colleagues Alexander Koplenig, Jan-Oliver Rüdiger and Sascha Wolfer for their discussions about research on gender-inclusive language. In addition to all of our appreciated colleagues in the field of gender linguistics, we would particularly like to thank Luise F. Pusch for her pioneering and tireless work in the field, from which we all have benefited, not only in this article.

[2] 'The women's rights activists may despair, but it cannot be changed: the language keeps it with the man. It is still not emancipated.' Kraus, Karl 1912: Die Abgeordnete. In: Die Fackel, 14. Jg., Heft 351-353 S. 66 (https://fackel.oeaw.ac.at/F/351,066), In: AAC - Austrian Academy Corpus: AAC-FACKEL Online Version: »Die Fackel. Herausgeber: Karl Kraus, Wien 1899-1936« AAC Digital Edition No 1.

## 1   Introduction

Shifting social norms are a driving force for language change, in particular regarding the debate around language and gender. In this debate, a recurring argument is that the change towards more gender-inclusive language has nothing to do with 'natural' language change: "… so the attempts of gender linguistics remain in the realm of re-labellings. They are being aggressively offered by a small group to the majority of the population against their will. They have nothing to do with language change whatsoever […]"³ (Bayer 2019a, cf. also 2019b). Eisenberg, a renown German linguist, finds similar words regarding the use of the so-called gender asterisk, a new orthographic symbol inserted between the masculine stem and the feminine suffix of personal nouns, e.g. in *Wissenschaftler\*innen* ('scientists'), which is intended to encompass all gender identities[4]: "The gender asterisk is a private invention and an ugly thing that tears up words and deforms texts, but its use also has unacceptable linguistic consequences"[5] (Eisenberg 2020: 9). Such comments give the impression that linguistic 'interventions' are unprecedented and not a natural part of language as a living system that is constantly changing. However, language change based on shifting social norms, as is the case with gender-inclusive language following the feminist and LGBTQ* movement, is neither unusual nor new. To illustrate this, we bring attention to another 'language battle' concerning the terms *schwul* ('gay') and *lesbisch* ('lesbian') which was being waged in the German Bundestag (section 2). In section 3, we trace the slow development of gender-inclusive language in the Bundestag regarding pair forms and feminized personal nouns . A discussion of the implications of these former language struggles for current issues of gender-inclusive language follows in section 4, and the symbolic value of linguistic changes will be discussed in section 5.[6]

---

³ Own translation, original: „Und daher bleiben die Versuche der Gender-Linguistik im Bereich von Umetikettierungen. Sie werden von einer kleinen Gruppe in aggressiver Weise der Bevölkerungsmajorität gegen deren Willen angedient. Mit Sprachwandel […] haben sie rein nichts zu tun."

---

[4] In German (as in other languages), new ways to address non-binary people more explicitly have been developed by the language community in recent years. An option already well established in the language system is to use neutralisations such as epicene nouns or derivatives of adjectives and verbs in the plural. New forms are an extension of established feminization strategies (pair forms *Lehrerinnen und Lehrer*, 'female and male teachers') with new gender symbols that are intended to explicitly include non-binary people (*Lehrer\*innen*, *Lehrer:innen*, 'teachers'). These symbols work particularly well in the plural, because German nouns and dependent elements of the noun phrase do not vary for gender in the plural, unlike in Italian or French, for example. Some qualitative studies have already found tendencies for less masculine generics and more gender-inclusive forms (cf. Adler & Plewnia 2019, Elmiger et al. 2017, Krome 2020). Quantitative studies on the use of these symbols are still scarce (e.g. Sökefeld 2021).

[5] Own translation, original: „Der Genderstern ist eine private Erfindung und ein hässliches Ding, das Wörter zerreißt und Texte verunstaltet, seine Verwendung hat jedoch auch unakzeptable sprachliche Konsequenzen."

## 2   'Sprache der Gosse' ('language of the gutter') in the German Bundestag

Tracing the history and successes of the gay and lesbian movement and the linguistic battles it has undertaken is outside the scope of this article. Suffice it to recall that it was in 1969 that the criminal prosecution of lesbians and gays was abolished in Germany, and only in 1990 did the WHO officially decide that homosexuality was not a mental illness. In Germany, the infamous paragraph §175 was removed from the constitution in 1994, and same-sex sexual acts were legally equated with heterosexual ones. The words *schwul* and *lesbisch*, similar to English *gay* and *lesbian*, had long been stigmatized, i.e. they were defamatory and insulting terms used to denote homosexual people. Traces of this could be found in youth language, where *schwul* and sometimes also *lesbisch* were typically used in a highly discriminatory and derogative way, for instance as a synonym for *revoltingly bad* or *unattractive*.[7] However, the lesbian and gay movement itself 'hijacked' these words and used them to refer to themselves, so that today they are often used neutrally (cf. Eitz 2010). This phenomenon of reappropriating pejorative terms can also be observed in other languages and lexical

---

[6] Parts of the first two chapters have already been published in German in Müller-Spitzer (2022).

[7] Duden states that its use is old-fashioned and not really found anymore in today's youth language. However, the connotation remains and a derogative usage is still possible, although we have not current data on the actual frequency and domain of use.

domains, "e.g. the word 'queen' to refer to an effeminate man" (Calder 2020: 2) in English.

An incident from the year 1988 in the German Bundestag illustrates language discussions around reappropriation and self-designation (for more details cf. Pusch 1994). The incident was triggered by a number of linguistic battles that went to court in the late 1980s. In 1988, the *Feministische Frauengesundheitszentrum Berlin* ('Feminist Women's Health Centre Berlin') wanted to issue an advertisement containing the word *lesbians*. The *Deutsche Postreklame GmbH* refused to print the ad, claiming that *lesbians* went against 'good morals'. The Women's Centre's objection was dismissed by the Frankfurt Regional Court, with a justification that sounds almost paradox to modern ears: Because of its vulgar choice of words, they claimed the text was disrespectful towards "those women who, in their erotic sensibilities, are attracted to female partners" (Pusch 1994: 248); in other words, it would be discriminating against lesbian women.

This conveys the impression that the local court prioritizes its judgment of which designation is appropriate in this context over that of a feminist women's centre . Four members of parliament from the party *Die Grünen* ('The Greens', since 1993 'Alliance 90/The Greens', is a green political party[8]) brought this linguistic dispute to the attention of the Bundestag by means of a *Kleine Anfrage* (literally: 'small request', i.e. a very condensed request addressed to the executive of the government), asking whether the federal government could use its influence to ensure that the above-mentioned advertisement was allowed to appear in Berlin, and whether the federal government was aware that lesbian support groups have the word *lesbisch* ('lesbian') in their name. The MPs also asked whether the federal government could guarantee "the right to self-designation in the sense of an emancipatory expression of opinion for gays and lesbians"[9] (Pusch 1994: 249–250).

---

[8] https://en.wikipedia.org/wiki/Alliance_90/The_Greens.

The President of the German Bundestag, Philipp Jenninger, rejected the inclusion of the request (the *Kleine Anfrage*) because of its wording, arguing that the terms *Schwule* and *Lesben* would not be accepted by the entire House. He would only include it in the agenda if *Homosexuelle und Lesbierinnen* ('homosexuals and lesbians'; the latter is an outdated term which more refers to female inhabitants of Lesbos[10]) were used. Similar to English, where the term *homosexual* "had medical and pathologizing connotations" (Calder 2020: 2) and was slowly replaced by the movement itself with the positive term *gay*, the terms demanded by the President were considered pejorative terms and not accepted by the gay and lesbian community. The applicants' written explanation – that *Schwule* and *Lesben* were the freely chosen self-designations and that the proposed alternatives were neither appropriate nor acceptable – was answered by Annemarie Renger, since Jenninger had meanwhile resigned. In her statement she said:

---

[9] Own translation, original: „Ist die Bundesregierung bereit, […] das Recht auf Selbstbezeichnung im Sinne einer emanzipatorischen Meinungsäußerung für Schwule und Lesben ggf. zu garantieren?"

[10] https://www.duden.de/rechtschreibung/Lesbierin_Frau.

> The terms 'gay and lesbian movement' may have passed over from colloquial to standard language in the meantime, but they still cannot be accepted by all members of the House. Let me remind you that on September 29, 1988, the vast majority of the Council of Elders also voted against allowing the use of such terms (cited in Pusch 1994: 253)[11].

CSU (*Christian Social Union*, formsa centre-right Christian democratic political alliance with the CDU, the *Christian Democratic-Union*[12]) member Wittmann stressed that the terms *Schwule* and *Lesben* were 'vocabulary belonging to the gutter' and not worthy to be used within the parliament (the *Hohe Haus*)[13] (Pusch 1994: 253).

The documentary film *Die Unbeugsamen* ('The Unbending') by Torsten Körner vividly shows how strong the everyday discrimination against gays still was at that time (for an overview on the history of gay people within the Federal Republic of Germany cf. Könne 2018). In the film (excerpt in figure 1), the CSU politician Hans Zehetmair (then Bavarian Minister of Education) points out in a talk show that in his view homosexuality is 'against nature' (*contra natura*) and 'a pathological behaviour' (*ein krankhaftes Verhalten*), and that society should spend less energy trying to 'understand the margins better' (*die Ränder besser zu verstehen*). Instead, the 'protection of the majority' (*Schutz der Vielen*) must be prioritized . 'Margins', however, 'should be thinned out' (*die Ränder müssen ausgedünnt werden*). Overt homophobia in political settings was therefore still socially acceptable, and few politicians acted against this widespread attitude (e.g. CDU politician and former minister Rita Süssmuth; interview following the talk show with Hans Zehetmair).

---

[11] Own translation, original: „Die Begriffe ‚Schwulen- und ‚Lesbenbewegung' mögen zwar inzwischen von der Umgangs- in die Hochsprache übergegangen sein, sie können aber trotzdem nicht von allen Mitgliedern des Hauses akzeptiert werden. Ich darf daran erinnern, daß sich auch der Ältestenrat am 29. September mit breiter Mehrheit dagegen ausgesprochen hat, die Verwendung derartiger Begriffe zuzulassen."

[12] https://en.wikipedia.org/wiki/CDU/CSU.

[13] Own translation, original: *Lesben und Schwule* seien "der Gosse zugehörige Vokabeln" und „des Hohen Hauses unwürdig".

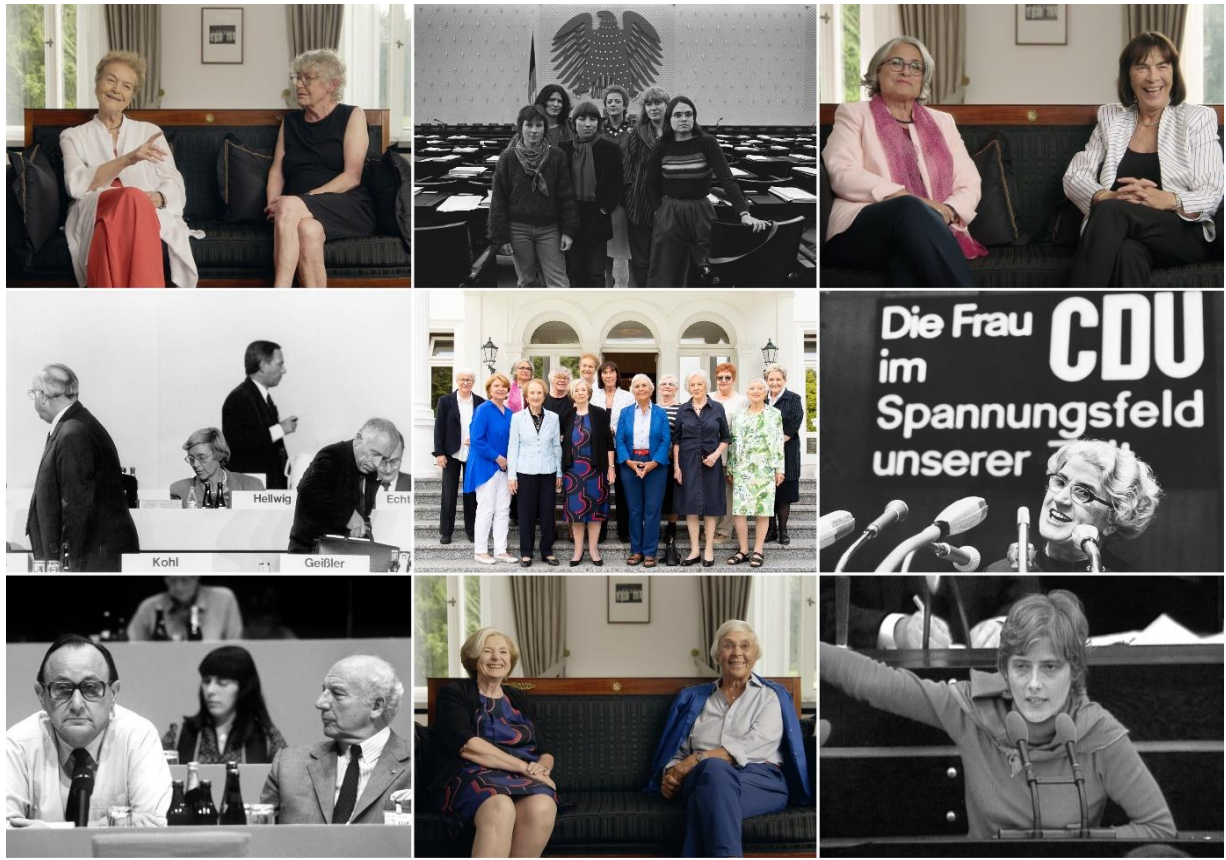

Fig. 1: Clip from the documentary "Die Unbeugsamen" (01:03:51-01:04:47; Copyright: Broadview Pictures; https://www.dieunbeugsamen-film.de/; 2020).

This also manifests itself in the further development of the linguistic discussion in the parliament. After the refusal to negotiate the application if it contained the terms *Schwule* and *Lesben*, the Green party switched from confrontational to subtle methods and reformulated the request with a made-up self-designation term found in a book from the 19th century (*Urninge und Urninden*[14]), which is how the motion was actually negotiated (cf. Fig. 2).[15]

**Deutscher Bundestag**
**11. Wahlperiode**

**Drucksache 11/3741**
15. 12. 88

**Antrag**

der Abgeordneten Frau Oesterle-Schwerin, Frau Kelly, Frau Olms, Volmer, Dr. Daniels (Regensburg), Häfner, Kreuzeder, Frau Rust, Frau Saibold, Weiss (München) und der Fraktion DIE GRÜNEN

Beeinträchtigung der Menschen- und Bürgerrechte der britischen Urninge und Urninden durch die Section 28 des Local Government Bill sowie vergleichbare Angriffe auf die Emanzipation der Urninge und Urninden in Bayern

Fig. 2: Excerpt from the protocol of the German Parliament (December 15, 1988 ), containing the terms Urninge and Urninden (https://dserver.bundestag.de/btd/11/037/1103741.pdf ).

In the explanatory note, the authors wrote that the terms stemmed from an 1864 writing on *mannmännliche Liebe* ('male-male love'), and that they preferred to fall back on this antiquated self-designation rather than use the outsider designations *Homosexuelle und Lesbierinnen*.[16] It was not until 1991 that the dispute was settled. The Greens tabled another request with the terms *Schwule* and *Lesben*, giving the Parliament the choice of accepting it as it was or, if rejected, publishing the rejection with the aim of maximum media coverage. The Parliament then accepted the request in the submitted form, so that since that time the self-designations *Schwule* and *Lesben* were allowed to be included in the official negotiations and thus in the protocols (Pusch 1994: 258).

---

[14] Karl Heinrich Ulrichs: „Vindex": social-juristisch Studien über mannmännliche Geschlechtsliebe; https://publikationen.ub.uni-frankfurt.de/opus4/frontdoor/deliver/index/docId/15279/file/Vindex-1F.pdf.

[15] Deutscher Bundestag, Drucksache 11/3741: https://dserver.bundestag.de/btd/11/037/1103741.pdf.

[16] Deutscher Bundestag, Drucksache 11/3741: https://dserver.bundestag.de/btd/11/037/1103741.pdf; ; p.2.

What is interesting about this linguistic battle is that the rejection of the terms by the CDU/CSU parliamentary group and parts of the SPD ('Social Democratic Party of Germany', a centre-left social democratic political party[17]) as well as the court decision against them, are not openly directed against the linguistic-emancipatory efforts of the gay and lesbian movement. Rather, the rejection of the terms is framed as protecting the gay and lesbian community, as it is assumed that outsiders are entitled to impose and demand more 'dignified' alternatives.[18]

We see the results of these linguistic struggles and discussions reflected in the corpus data of the German Bundestag plenaries. Figure 3 shows the distribution of the terms *homosexuell, schwul, lesbisch* and *queer* in the Bundestag protocols from the 2nd to the 19th legislature (from 1953 to 2021), which we extracted via regular expressions from the Corpus of Protocols of the German Bundestag (Müller & Stegmeier 2021). The topic of homosexuality is barely discussed at all before the 9th legislature (1980-1983), after which the term *homosexuell* is most prominent for several years. It is only in the 11th legislature (1987-1990), that *schwul* and *lesbisch* start to appear regularly in the

---

[17] https://en.wikipedia.org/wiki/Social_Democratic_Party_of_Germany.

[18] For moral aspects of using/not using self-designations, cf. Stefanowitsch 2018: 55–60; for sociological perspectives on self- vs. outsider-categorization and the interactions between both, cf. Hirschauer 2021: 164–165.

protocols, most likely due to the discussions initiated and furthered by the Greens in 1988. From then on, the two terms appear almost always equally. This can be explained by the frequent use in pair forms, e.g. *schwule und lesbische Paare* ('gay and lesbian couples'), *lesbische Frauen und schwule Männer* ('lesbian women and gay men'), and *Schwule und Lesben* ('gays and lesbians'). The hypernym *homosexuell* remains frequent throughout the decades, and only in the 17th legislature is it less frequent than *lesbisch* and *schwul*. This demonstrates that even though lesbian and gay communities chose *schwul* and *lesbisch* as self-designations, the use of the umbrella term *homosexuell*, to denote both, is often dominant. Contrary to English, where *gay* can be used as a synonym for *homosexual* (cf. Obama's inauguration speech further down), the term *schwul* in German is completely restricted to male homosexuals. Therefore, the existence and use of the underspecified hypernym *homosexuell* is still necessary in German (or, as another strategy, the use of both *schwul* and *lesbisch* in pair forms). In the 16th legislature (2005-2009), we see the emergence of the new umbrella term *queer*, an English loanword that comprises much more than *homosexuell* (cf. Baker 2013). It is "a move beyond looking at dominant categories as homogenous identities towards a more inclusive understanding of non-normative sexuality" and gender identity (Jones 2021: 15). These conceptual changes go hand in hand with cultural frame shifts, i.e., "a sexual preference develops from the sin of sodomy to a medicalized concept of 'homosexuality', and then to a private lifestyle of same-sex romantic relationships with the option of an indiscriminate 'marriage for all'."[19] (Hirschauer 2020: 329)

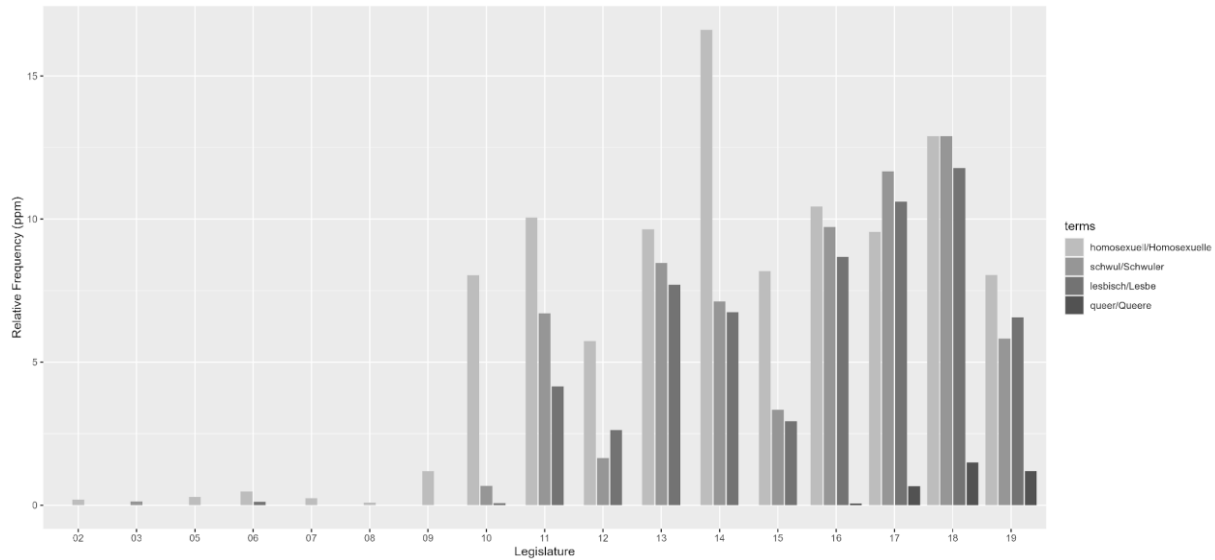

Fig. 3: Relative frequencies of the adjectives *homosexuell*, *schwul*, *lesbisch*, *queer* and their nominal derivations in Bundestag protocols from 1953 to 2021.

The battle to increase social visibility and recognition of non-heterosexual relationships has been waged for a long time, and continues still. According to a 2017 survey conducted by the Federal Anti-Discrimination Agency, 44% of respondents said that homosexuals should stop "making such a fuss about their sexuality" (Küpper et al. 2017). On the other hand, many fundamental things have been improved in Germany, such as the legal equality of homosexual and heterosexual couples, following the legal innovation of the *Ehe für alle* ('marriage for all').[20] Internationally, many considered it a linguistic and political milestone when Barack Obama used the word *gay* for the first time in the inaugural address of his second term. Referring to the American Declaration of Independence, he declared:

> It is now our generation's task to carry on what those pioneers began. For our journey is not complete until our wives, our mothers, and daughters can earn a living equal to their efforts. Our journey is not complete until our gay brothers and sisters are treated like anyone else under the law — for if we are truly created equal, then surely the love we commit to one another must be equal as well.[21]

---

[19] Own translation. Original: „wenn etwa eine sexuelle Präferenz sich von der Sünde der Sodomie über die medikalisierte ‚Homosexualität' zum privaten Lebensstil der geschlechtsgleichen Paarbeziehung mit der Option auf differenzlose ‚Ehe für alle' entwickelt"

[20] Cf. https://www.lsvd.de/de/ct/934-Von-1933-bis-heute-Lesben-und-Schwule-in-Deutschland-und-der-DDR.

However, this was immediately interpreted as a "declaration of war on conservative America" (Kolb 2013), and thus shows parallels to the language battles in the Bundestag, where a certain language use was associated with certain social positions and allegiances. However, more recent developments in the US-American context show a reactionary backlash against these liberal moves (cf. Hirschauer's notion of "regenderings" in times of "degenderization" (2020: 330)), e.g. the 'Don't say gay' bill in Florida[22] as well as severe restrictions of abortion rights[23].

---

[21] https://www.nj.com/politics/2013/01/presidential_inauguration_2013.html.

[22] https://edition.cnn.com/2022/03/28/politics/dont-say-gay-bill-desantis-signs/index.html

[23] https://www.guttmacher.org/2023/01/six-months-post-roe-24-us-states-have-banned-abortion-or-are-likely-do-so-roundup

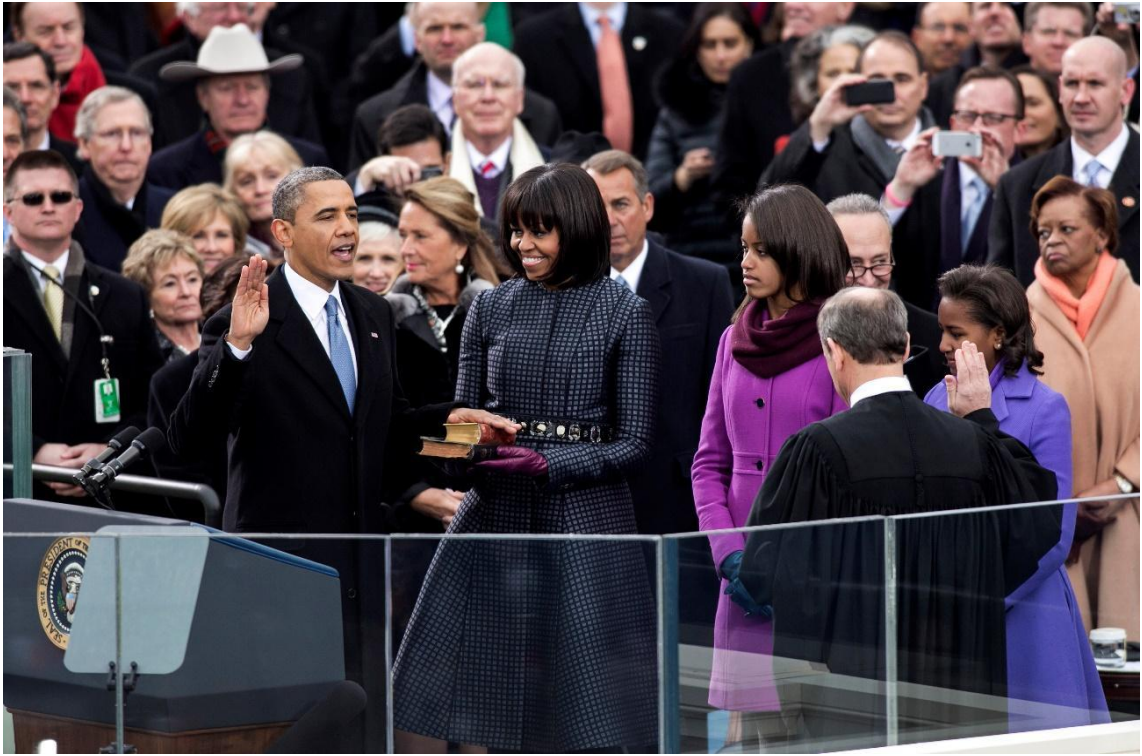

Fig. 4: President Barack Obama takes the oath of office from Supreme Court Chief Justice John G. Roberts Jr., right, in a public ceremony at the U.S. Capitol before thousands of people in Washington, D.C., Jan. 21, 2013. Roberts administered the oath in an official ceremony at the White House, Jan., 20, 2013
(https://en.wikipedia.org/wiki/Second_inauguration_of_Barack_Obama#/media/File:Barack_Obama_second_swearing_in_ceremony_2013-01-21.jpg).

An important step that the LGBTQ* movement has long called for in Germany is an amendment to the Constitution. In 2019, a passage was added to the *Grundgesetz* that no one may be discriminated against on the basis of their "sexual identity".[24] In addition, the federal government commissioned a *Queer-Beauftragten* ('queer representative') for the first time in 2022.[25]

---

[24] Deutscher Bundestag, Drucksache 19/13123: https://dserver.bundestag.de/btd/19/131/1913123.pdf.

[25] https://www.bmfsfj.de/bmfsfj/ministerium/behoerden-beauftragte-beiraete-gremien/queer-beauftragter-der-bundesregierung.

It is clear from these shifts in language use and underlying social norms that these two factors are intrinsically intertwined. Who claims a voice and who listens to it? Who claims to know what is the 'appropriate' term for a group? We can also see that these battles are often fought under false pretences, and that the real motives for rejecting the linguistic visibility of gay people were being deliberately masked. We also experience that self-designations can be continuously changing, and that the linguistic community is in a constant process of negotiating these terms, e.g. in the English-speaking world, where explicit labellings of 'gay' and 'lesbian' seem to be on the decline in favour of more inclusive and less categorical terms like 'queer' and 'LGBT' (Baker 2013, Jones 2021: 15).

## 3 The slow development towards more gender-inclusive language in the German parliament

Just as the slow but steady changes in the acceptance of queer self-denominations, current efforts regarding gender-inclusive language in German (cf. e.g. Lind & Nübling 2022, Müller-Spitzer 2022 und 2021, Simon 2022) are embedded into long-standing feminist emancipation efforts. Three examples from the Bundestag are presented to illustrate this.

### 3.1 *Frau* (Ms.) but not *Herr* (Mr.): Female forms of address in the list of deputies

Deviations from a presumed norm tend to be marked linguistically. This has been discussed in the context of androcentrism, or *male as norm* (MAN; cf. Bailey et al. 2019), which manifests itself linguistically, e.g. in the formation of compounds like *Frauenfußball* or *Frauenmannschaft* ('women's football', 'women's team') as opposed to uncompounded *Fußball* or *Mannschaft* ('football', 'team') which usually refer to male domains if not specified otherwise (cf. Kotthoff & Nübling 2018: 135, Pusch 1984a: 98, 100–101, Trömel-Plötz 1978: 57). In few cases, we find *female as norm*, with the male being marked as the deviation (e.g. *Parfüm* vs. *Herrenparfüm*; 'perfume' vs. 'men's perfume') (cf. Hornscheidt 2008 for a discussion on deviations in Swedish). Other linguistic markings of deviations from supposed norms are found in racist terminologies, e.g. using *white* and *coloured* in direct opposition – as if light-skinned people do not also have a skin colour or 'race'. As such, the Black and PoC communities, just as the queer community, are in a constant process of finding positive self-denominations, for instance *Black* (with a capital B) or *Person/People of Colour* (PoC). Still, white linguistic supremacy and Anti-Black Linguistic Racism are prevailing problems, stemming from *white as norm* concepts (cf. Baker-Bell 2020).

The Bundestag protocols are another example of how deviations from the norm (in this case from MAN) were marked linguistically for a long time. In the name lists of the plenary protocols, the address *Frau* ('Ms.') was added to the names of female deputies. For male deputies, only the surnames were listed.[26] The Bundestag name lists then read, for example: "Dr. Ahrens, Baum, **Frau** Beer, Dr. Biedenkopf, Biehle, Büchner (Speyer), Carstensen (Nordstrand), **Frau** Dr. Däubler-Gmelin"[27] (cf. Fig. 5, also Fig. 2).

---

[26] It has been shown that men are more often referred to by surname only, whereas first and last name or the address *Frau* tend to be used for women; cf. Bühlmann 2002: 185; for English, cf. Atir & Ferguson 2018, McConnell-Ginet 2003.

[27] Deutscher Bundestag, Plenarprotokoll 11/96, 6565, https://dserver.bundestag.de/btp/11/11096.pdf.

```
Liste der entschuldigten Abgeordneten

Abgeordnete(r)              entschuldigt bis einschließlich

Dr. Ahrens*                  29. 9.
Baum                         30. 9.
Frau Beer                    30. 9.
Dr. Biedenkopf               30. 9.
Biehle                       30. 9.
Büchner (Speyer) *           28. 9.
Carstensen (Nordstrand)      30. 9.
Frau Dr. Däubler-Gmelin      30. 9.
Frau Dempwolf                30. 9.
Ehrbar                       30. 9.
```

Fig. 5: Excerpt from the list of deputies of the German Bundestag from 1988 (Deutscher Bundestag, Plenarprotokoll 11/96, p. 6565, https://dserver.bundestag.de/btp/11/11096.pdf).

It was not until 1991 that this practice was abandoned, but not without much debate. In 1987, for example, Marliese Dobberthien of the SPD criticised the exclusive use of the female form of address in Bundestag protocols:

> All print matter of the Bundestag also contains completely unnecessary gender-specific formulations in which men are the standard, the norm, but women are only the exception to which special attention must be paid. In registers, for example: Why don't we put a *Mr.* in front of the male deputies? Don't our men deserve a little more courtesy?[28]

The first women in the Bundestag were exceptional in the political arena, and this exceptionality was linguistically highlighted. It was only 30 years ago that the gender-specifying form of address and thereby the exceptional status of women was removed from the protocols.

---

[28] Own translation, original: „Auch alle Bundestagsdrucksachen weisen völlig unnötige geschlechtsspezifische Formulierungen auf, wo Männer die Regel, die Norm sind, die Frauen aber nur das Besondere, auf das man eigens hinweisen muß. Zum Beispiel in Verzeichnissen: Warum setzen wir nicht auch ein „Herr" vor die Abgeordneten bei Auflistungen? Haben unsere Männer nicht ein bißchen mehr Höflichkeit verdient?"; Deutscher Bundestag, Plenarprotokoll 11/37, p. 2503, https://dserver.bundestag.de/btp/11/11037.pdf.

## 3.2 Feminization: *Frau Präsident* or *Frau Präsidentin*?

According to Hellinger & Bußmann (2003: 150–160), when referring to persons in German, we can distinguish personal nouns that specify referential gender by grammatical, lexical or morphological means. This leads to different kinds of pair forms: a) lexical pairs (*Krankenschwester/Krankenpfleger* 'nurse/male nurse'; *Vater/Mutter* 'father/mother'); b) double gender nouns (nominalized adjectives and participles in the singular), i.e. the grammatical gender of the article and the adjective inflection are the only overt difference between masculine and feminine forms (e.g. *der Abgeordnete [m.]/die Abgeordnete [f.]*, 'deputy'). This distinction is neutralized in the plural, since articles and other determiners do not vary for grammatical gender in the plural (*die Abgeordneten*, 'the deputies'); c) most prominently, masculine forms for which there are feminine derivations, usually realized with the suffix *-in* (e.g. *Minister/Ministerin*, 'minister'; *Präsident/Präsidentin*, 'president'). These word pairs are semantic minimal pairs in that they have the opposing semantic features *+male/-female* and *+female/-male* (Diewald 2018: 290–293). Within these minimal pairs, the masculine form usually has two functions: first, as a masculine specific, and second, as a so-called 'masculine generic'. The term denotes the use of the masculine form to refer to a group of people whose gender is unknown, irrelevant or ignored (e.g. *Wissenschaftler*, 'scientists', for a group of scientists) and is used traditionally in a lot of natural and grammatical gender languages to refer to people in a generic way (Hellinger & Bußmann 2001). However, it is unusual to use the masculine form to refer to an individual woman because singular specific expressions (especially direct addresses) are highly referential, leading to the expectation that the grammatical gender indicates referent gender, i.e. a masculine singular indicates a male individual, a feminine singular indicates a female individual (Kotthoff & Nübling 2018: 93). Becker even calls female-referring masculines a "lie" (2008: 66), e.g. if meeting with a female colleague was announced by the utterance *Heute Abend gehe ich mit einem Kollegen zum Essen* ('Tonight I'll have dinner with a [male] colleague'). In actual language use, however, we find instances in which the masculine is used to refer to specific women, see.g. in the protocols of the German parliament. When Annemarie Renger was the first woman to be elected President of the Bundestag in 1972, the Senior President (*Alterspräsident*) addressed her as *Frau Präsident* (masculine form of 'president')[29] in his congratulations. Other female ministers such as Katharina Focke (1972-1976), were also addressed with the male form, e.g. as *Minister*: *Focke, Minister für Jugend, Familie und Gesundheit* ('Focke, Minister[masc.] for Youth, Family and Health'). This practice continued until the end of the 1980s, even though debates were hinting at a coming change, for example, protests from deputies when the masculine form was used to refer to a woman.[30]

Figure 6 shows our analysis of the forms *Frau Präsident* vs. *Frau Präsidentin* in the corpus data of the German Bundestag plenaries. *Frau Präsident* (female address +

---

[29] „Frau Präsident, ich übermittle Ihnen die Wünsche des Hauses und bitte Sie, diesen Platz einzunehmen."; Deutscher Bundestag, 1. Sitzung, 13. Dezember 1972, S. 3, https://dip21.bundestag.de/dip21/btp/07/07001.pdf.

[30] „Vizepräsident Cronenberg: Das Wort hat der Bundesminister für Jugend, Familie, Frauen und Gesundheit, Frau Süssmuth. (Frau Schmidt [Nürnberg] [SPD]: Das ist die Bundesministerin!) — Frau Ministerin, Sie haben das Wort." Deutscher Bundestag, Plenarprotokoll 11/96, 6559, https://dserver.bundestag.de/btp/11/11096.pdf.

masculine form) existed for some time (6th-11th legislature) but was never more frequent than *Frau Präsidentin* (female address + feminine form).[31] After the 11th legislature, it disappeared almost completely, until in the 19th legislature, there was a very slight rise again. This is mostly due to the AfD ('Alternative for Germany', a right-wing populist party[32]), contributing 40 of the 49 usages (81.6%) in this legislature. However, we see that the decision to reinforce the use of feminine forms for female referents had the desired effects. If a party counteracts such regulations, it is often a conscious reactionary act that linguistically indexes the party line (in the case of AfD, their rigorous rejection of gender-inclusive language and their populist fight against it, as well as their anti-feminist and anti-queer agenda[33]).

---

[31] It is important to note that these addresses of (female) presidents are not limited to presidents of the Bundestag (besides Annemarie Renger, there were Rita Süßmuth in the 12th and 13th legislature and Bärbel Bas in the current, i.e. 20th legislature, which is not yet part of the dataset) but also include other presidents, e.g. *Alterpräsidentin Frau Dr. Dr. h. c. Lüders* (oldest member of parliament, leading the first session until a regular president is elected) in the 2nd legislature, or female vice-presidents (e.g. Anke Fuchs in the 14th legislature or Edelgard Bulmahn in the 18th).

[32] https://en.wikipedia.org/wiki/Alternative_for_Germany.

[33] Cf. a recent debate in the Bundestag on gender-inclusive language (https://www.bundestag.de/mediathek?videoid=7555395#url=L21lZGlhdGhla292ZXJsYXk/dmlkZW9p ZD03NTU1Mzk1&mod=mediathek): The AfD is the only party on the federal level that repeatedly demands bans on gender-inclusive language. They have made five motions in the past years, while other parties have not motioned for gender-inclusive language as a political topic at all. This leads to the (superficially) paradoxical situation in which the party that fights most against it also brings the topic up time and time again. The reactionary role of right-wing parties in the contestations of gender equality and gender-inclusive language has been observed on a global level (cf. Erdocia 2022, Roth and Sauer 2022).

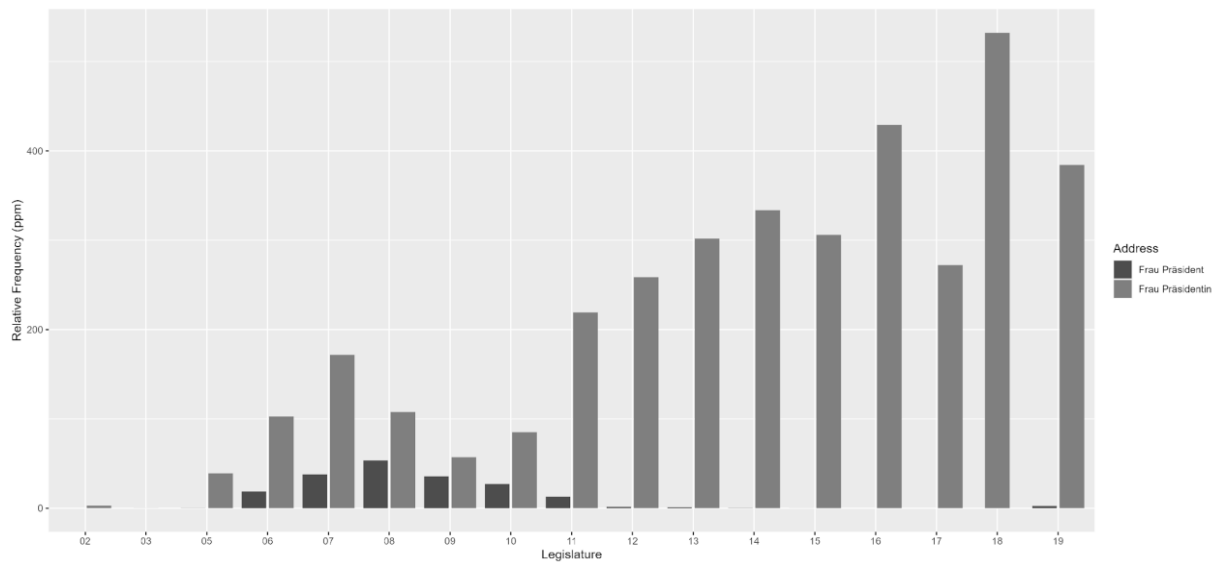

Fig. 6: Relative frequencies of the addresses *Frau Präsident* and *Frau Präsidentin* in Bundestag protocols from 1953 to 2021.

From the 1980s onwards, we generally see "a dramatic increase" in the use of female forms, which had been "virtually non-existent in the debates before" (Stecker et al. 2021: 1). Today, the *Protokoll Inland* ('Domestic Protocol') states that female officials should be addressed with feminized forms (e.g. *Präsidentin* or *Ministerin*) both orally and in writing (cf. Bundesministerium des Innern 2016; Wissenschaftliche Dienste Deutscher Bundestag 2021). The AfD is the only party that has repeatedly broken this practice (cf. Hallik 2020: 82–84, cf. also Fig. 6).

The naming practice was officially changed in the course of a resolution on gender-fair legal language, a project that was considered necessary by all parliamentary groups in the early 1990s. However, many legal texts still remain in the generic masculine – inter alia article 69/3 in the constitution, where only the masculine forms of *president, chancellor*, and *minister* are found. In 2004, before the election of Angela Merkel as the first female chancellor, Schewe-Gerigk expressed her hope that this wording would change with a female chancellor or president (2004: 330). However, even after 16 years under the chancellorship of Angela Merkel, this legal text remains unchanged.[34]

### 3.3 Gender-inclusive legal language?

The following resolution, passed in the Bundestag on July 24, 1991 with the votes of the CDU/CSU and the FDP ('Free Democratic Party', a liberal political party[35]), reminds us of current debates about gender-inclusive language:

> The Federal Government is called upon to avoid gender-specific terms in all draft laws, ordinances and administrative regulations with immediate effect, and to either choose gender-neutral formulations or to use formulations that refer to both genders.[36]

This resolution was adopted on the basis of an application submitted by the CDU/CSU parliamentary group. The SPD, the Greens and the Left Party had originally tabled a more far-reaching resolution that would have imposed even more obligations, so they abstained from the vote. However, there was cross-party agreement that the exclusive use of masculine forms in legal language was no longer appropriate. This, too, had been preceded by long discussions, as Parliamentary State Secretary Rainer Funke (FDP) explained:

> This critical attitude towards our language is now accepted and taken seriously. This was not always the case. It was a long and arduous way to get there; a way that began with controversial

---

[34] https://www.gesetze-im-internet.de/gg/art_69.html

[35] https://en.wikipedia.org/wiki/Free_Democratic_Party_(Germany).

[36] Own translation, original: „Die Bundesregierung wird aufgefordert, ab sofort in allen Gesetzentwürfen, Rechtsverordnungen und Verwaltungsvorschriften geschlechtsspezifische Benennungen/Bezeichnungen zu vermeiden und entweder geschlechtsneutrale Formulierungen zu wählen oder solche zu verwenden, die beide Geschlechter benennen." Cf. Also Deutscher Bundestag 12/1041: Unterrichtung durch die Bundesregierung: Maskuline und feminine Personenbezeichnungen in der Rechtssprache, 07.08.1991: https://dserver.bundestag.de/btd/12/010/1201041.pdf.

> opinions, sometimes with mutual accusations by men and women, with exaggerated ideas and unobjective objections.[37]

The interconnectedness of language and gender equality was also made clear in the resolution:

> The correct way of addressing and referring to women is of great importance for the equal treatment of women and men in social reality. This particularly applies to official language that refers to specific facts and persons. But also the choice of words in regulations needs to be revised.[38]

Efforts to achieve equality on this linguistic level were closely linked to the successes of the women's movement. For example, the first resolution on the topic of gender-inclusive legal language was issued by the *Deutsche Frauenrat* ('German Women's Council') in 1982, in which they officially demanded the legislature use gender-inclusive language. This again was motivated by feminist linguists from the New Women's movement in the 1970s (Schewe-Gerigk 2004: 322). We see that all these political achievements did not come about just like that, but were fought for in every detail and in many discussions, especially by feminist groups.

---

[37] Own translation, original: „Diese kritische Haltung unserer Sprache gegenüber wird inzwischen ernst genommen und akzeptiert. Das war nicht immer so. Bis hierhin war es ein mühsamer und ein langer Weg; ein Weg, der mit kontroversen Auffassungen, zuweilen auch mit gegenseitigen Vorwürfen von Frauen und Männern, mit überspannten Vorstellungen und unsachlichen Repliken begann." Deutscher Bundestag, Plenarprotokoll 12/132, 11525, https://dserver.bundestag.de/btp/12/12132.pdf.

[38] Own translation, original: „Die korrekte Anrede und Bezeichnung von Frauen hat große Bedeutung für die Gleichbehandlung von Frauen und Männern in der sozialen Wirklichkeit. Dies gilt insbesondere für die auf konkrete Sachverhalte und Personen bezogene Amtssprache. Aber auch die Wortwahl der Vorschriften bedarf einer Überprüfung." Unterrichtung durch die Bundesregierung. Maskuline und feminine Personenbezeichnungen in der Rechtssprache, Drucksache 12/1041, https://dserver.bundestag.de/btd/12/010/1201041.pdf.

Since then, many new or revised laws have been formulated along these lines (old ones remain unchanged).[39] The generic masculine, such as the use of words like *Arbeitnehmer* ('employee') or *Minister* ('minister') for all people who fulfil this function, has continued to decline in legal and political language. This can be seen in the New Year's and Christmas addresses of the Federal Chancellors and Presidents of the last 30 years. The generic masculine has been replaced by binomials such as *Bürgerinnen und Bürger* ('female and male citizens') or *Polizistinnen und Polizisten* ('female and male police officers'), gender-neutral forms (epicenes) such as *Rettungskräfte* ('rescue workers'), *Alte* ('the elderly') or *Arbeitslose* ('the unemployed'), as well as pronominal paraphrases such as *wir alle* ('all of us') or *alle, die* ('all those who'). In his first New Year's address, the current chancellor Olaf Scholz (SPD) did not use a single masculine generic form (Müller-Spitzer et al. 2022). As binomials, epicenes and paraphrases are considered subtle and long-established forms of gender-inclusive language (cf. acceptability survey about different gender-inclusive forms, WDR 2023). Headlines such as *Olaf Scholz gendert* ('Olaf Scholz uses gender-inclusive language') did not appear. Had he used a glottal stop (the way in which new gender symbols such as the asterisk or the colon are phonetically realised, cf. Völkening 2022), this would have surely been a topic of debate.

---

[39] Here, e.g., examples from actual laws: „Beschäftigte im Sinne dieses Gesetzes sind 1. *Arbeitnehmerinnen und Arbeitnehmer*, …" (PflegeZG, §7, Absatz 1, 2021), „*Die* oder *der Datenschutzbeauftragte* wird auf der Grundlage ihrer oder seiner beruflichen Qualifikation …" (BDSG, Kapitel 3, §5, Absatz 2, 2021) [own highlights].

The use of binomials as unobtrusive forms of gender-inclusive language[40] has also become established in the Bundestag protocols. Figure 7 (our analyses are again based on the corpus by Müller & Stegmeier 2021) shows the constant rise of these forms in the plural, especially with the feminine form in the first place *(Kolleginnen und Kollegen*, 'female and male colleagues'). Binomials with the masculine form in the first slot (*Kollegen und Kolleginnen*, 'male and female colleagues') are much rarer. This 'feminine-first rule' in binomials has been observed in other studies as well (Rosar 2022; Truan 2019) and explained in various ways, for instance as a result of the efforts and successes of feminist linguistics and ensuing legal regulations (Rosar 2022: 287); more convincingly, regarding the referential properties of the forms: the more specific feminine form is placed first in order to prevent the masculine form from being understood generically (Eisenberg 2022). In an exemplary study of the binomial *Bürgerinnen und Bürger* ('female and male citizens') in plenary protocols, Müller also finds that it has been on the rise since 1979, while generic uses of the masculine form *Bürger* are declining (2022: 126)[41]. The same is true for New Year's and Christmas addresses. Binomials are on the rise, whereas masculine generics are starting to disappear (Müller-Spitzer et al. to appear). All of these results are in line with Truan's findings that "[i]n German political discourse specifically, it has become usual – or politically correct – to use [binomials]" (2019: 206).

---

[40] When asked about their attitudes towards gender-inclusive language by the German public-broadcasting institution WDR 2023 (https://www1.wdr.de/nachrichten/gender-umfrage-infratest-dimap-100.html), two thirds of the participants answered that they accept binomials as gender-inclusive forms (whereas only one third thinks that the new gender symbols are acceptable).

[41] However, Müller notes a slight rise of the masculine generic since 2018 due to the right-wing populist party AfD.

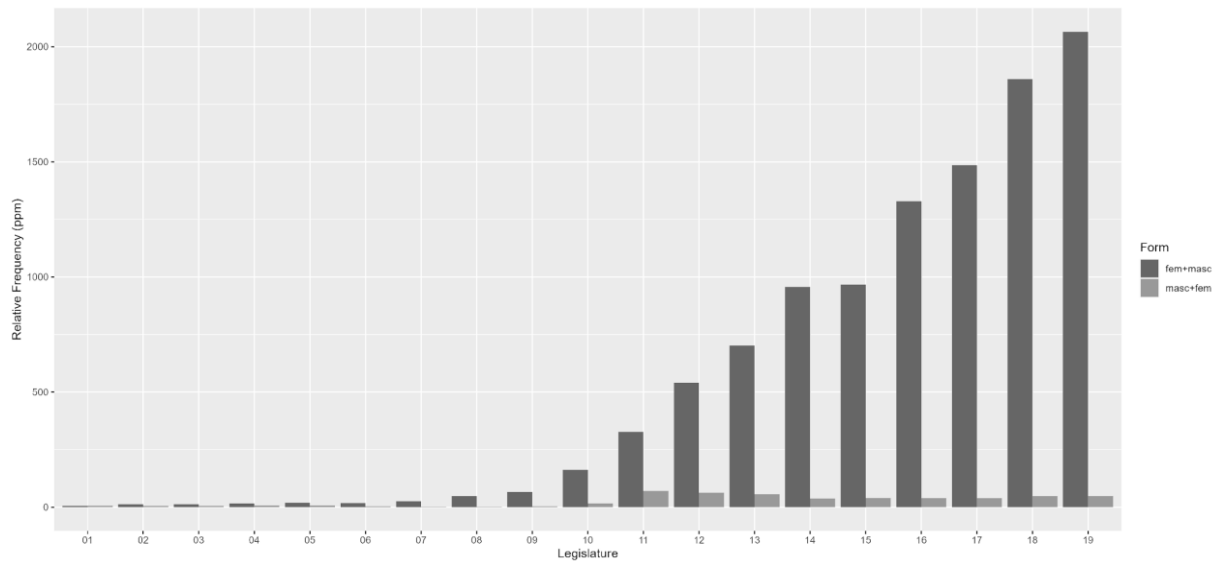

Fig. 7: Relative frequencies of binomials (feminine first, masculine first) in Bundestag protocols from 1949 to 2021. Only plural forms with the same lexical base were extracted (i.e. asymmetrical forms like *Damen und Herren* 'ladies and gentlemen' are not part of the analysis).

Figures 8 and 9 show the ten most frequent plural binomials for feminine and masculine first, respectively. We see that the direct address of colleagues in the parliament (*Kolleginnen and Kollegen, Kollegen und Kolleginnen*) is most prominent in both charts, followed by the 'topics' of political debates – the citizens (*Bürgerinnen und Bürger, Bürger und Bürgerinnen*). Most top-ten binomials are shared by both formats; only three forms are unique. In the feminine-first forms, we find *Patientinnen und Patienten* ('female and male patients'; 11[th] with the masculine form first). In the masculine-first forms, we see *Bauern und Bäuerinnen* ('male and female farmers') and *Rentner und Rentnerinnen* ('male and female pensioners') in 6[th] and 7[th] place, respectively. However, more data is needed to study these different sequences more systematically.

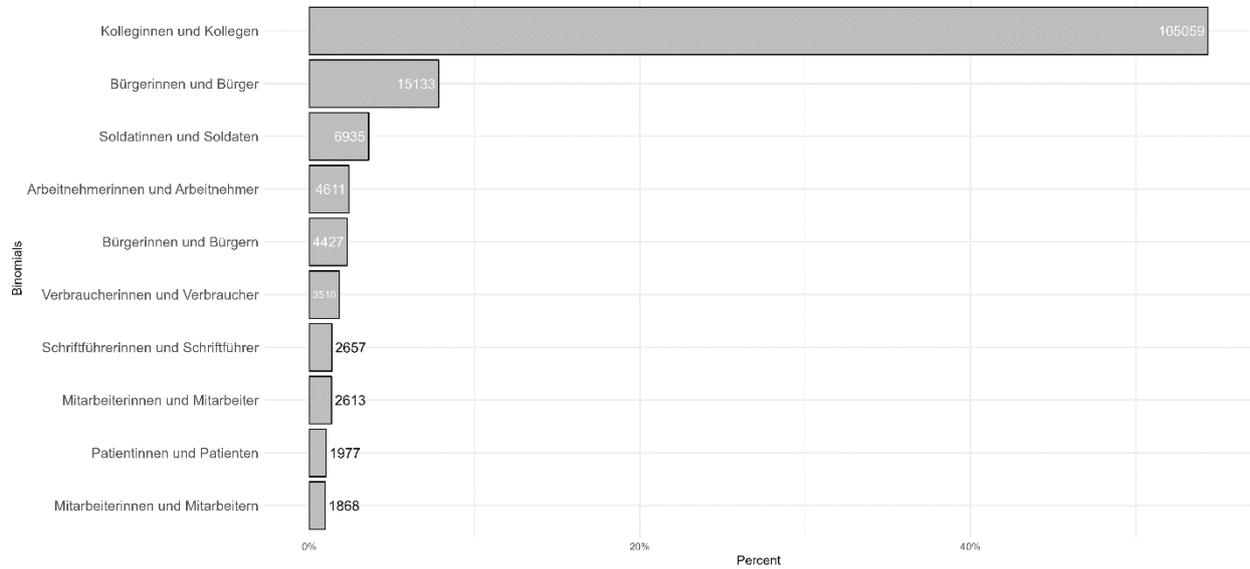

Fig. 8: The ten most frequent binomials with a feminine form in the first slot (total amount of types: 3,526; total amount of tokens: 193,242); Bundestag protocols 1949-2021.

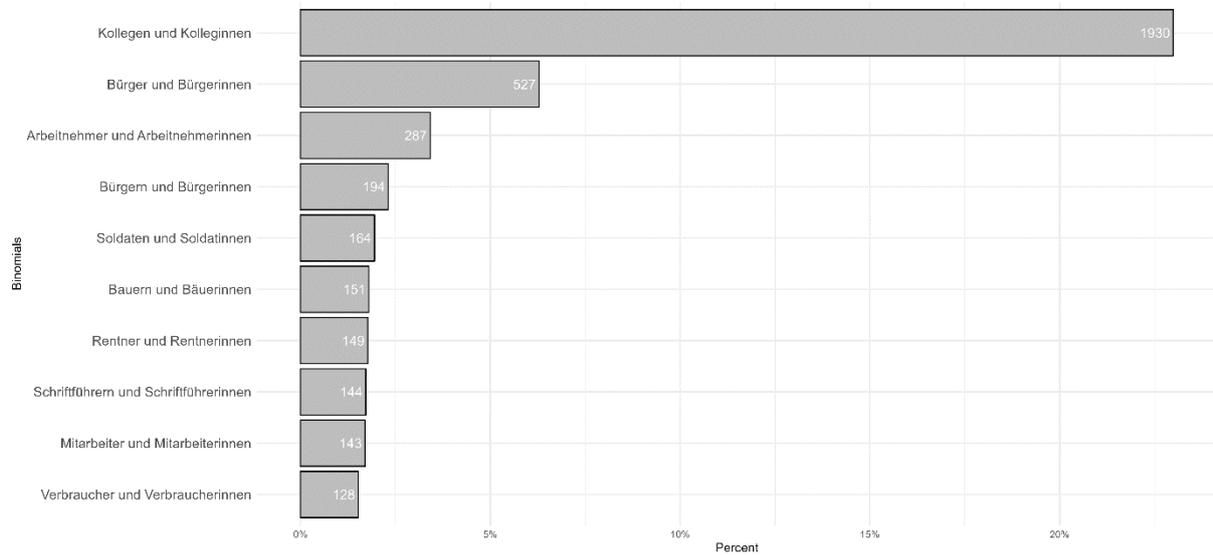

Fig. 9: The ten most frequent binomials with a masculine form in the first slot (total amount of types: 1,707; total amount of tokens: 8,396), Bundestag protocols 1949-2021.

Although much has changed in legal and everyday language, and the topic has been around for at least 50 years, the debate about new forms of gender-inclusive language (e.g. gender symbols) is particularly heated at present. In Germany, hardly a month goes by without a new survey, news program or talk show on the subject. Also linguists argue about the topic and its scientific dimensions (Eisenberg 2022, Müller-Spitzer 2021 and 2022, Simon 2022, Stefanowitsch 2018, Trutkowski & Weiß 2023, Zifonun 2018, and many more). It has gone so far that some federal states (e.g. Saxony, Saxony-Anhalt, Hesse, Bavaria) have started banning the use of gender-inclusive language in all official contexts (Lischewski 2023), while other federal states are in favour of raising this issue in the context of school education without prescribing or prohibiting the use of gender-inclusive language. Opponents of gender-inclusive language frequently claim that proponents want to actively impose the use of certain linguistic forms on society as a whole, however, cases like the above named prohibitions suggest that it is the other way around; rather, it is the opponents of gender-inclusive language who issue prescriptions and regulations regarding language use (Stefanowitsch 2018: 20).

## 4   Current debates about language and gender

The decline of the generic masculine in addresses, job titles, legal texts and many other contexts (cf. some qualitative studies on the topic, e.g. Adler & Plewnia 2019, Elmiger et al. 2017, Krome 2020; for a quantitative analysis of New Year's and Christmas addresses cf. Müller-Spitzer et al. to appear) is an ongoing process that was initiated by emancipation efforts of the women's movement and feminist linguists like Selma

Trömel-Plötz (1978) and Luise F. Pusch (1984b, 1982 and 1979). It has been reflected in politically motivated linguistic changes for more than 30 years now (cf. the use of feminine forms and binomials discussed above). These developments are by no means limited to the German-speaking world. Especially for English, a wide range of studies has addressed the use of the epicene pronoun *they* (also called *singular they*) and the ensuing decrease of the generic masculine *he* (Adami 2009, Balhorn 2004, Baranowski 2002, Meyers 1990, Paterson 2011, and many more). As Earp concludes, the use of masculine generics has "fallen dramatically" over time (2012: 4).

Today, some (academic) style guides encourage the use of singular *they* and ask authors not to use generic *he* (or *she*)[42], or they at least give authors the option to use it, although they would rather recommend to avoid it by rephrasing[43]. Similar developments can be observed in German language style guides and manuals regarding the use of gender-inclusive German, particularly in the context of universities, companies and state institutions. The diversity and plurality of available forms is reflected in the widely differing recommendations and guidelines, as well as in the variable use of such forms, for example, on city websites (Müller-Spitzer & Ochs 2023). In general, less disputed forms like binomials and neutralizations tend to be recommended instead of more salient strategies like gender symbols (at least in Switzerland, cf. Siegenthaler 2021: 85).

---

[42] e.g. APA (https://apastyle.apa.org/style-grammar-guidelines/grammar/singular-they).

[43] e.g. MLA (https://style.mla.org/using-singular-they/) and CMOS (https://www.chicagomanualofstyle.org/qanda/data/faq/topics/Pronouns/faq0031.html).

Interestingly, although gender symbols are neither prescribed nor necessarily recommended, and are only used in a limited amount of written (press) sources (cf. Waldendorf 2023)[44], they are the main focus of many current language 'battles' in Germany. The notion that the grammatical gender of personal nouns has nothing to do with the gender identity of the people they refer to (e.g. Eisenberg 2022, Kalverkämper 1979, Trutkowski & Weiß 2023) is already outdated, as can be seen in the debate about and the ensuing enforcement of the use of feminine forms to refer to female parliamentarians some 30 years ago. It also overlooks "decades of linguistic-semantic theory-building and empirical research on the question of how exactly linguistic signs convey meaning"[45]. Regarding masculine generics, several psycholinguistic studies have shown that they evoke a male bias rather than being understood neutrally (i.a. Gabriel et al. 2008, Körner et al. 2022, Zacharski & Ferstl 2023a), again highlighting that grammatical gender is thoroughly intertwined with referent gender and gender distributions in the real world (e.g. in jobs). Klein (2022: 179) has shown that even so-called epicene nouns (i.e. nouns that have a fixed grammatical gender but do not specify referent gender, e.g. *die Person* (f.) 'the person', *der Mensch* (m.) 'the human being') tend to be interpreted as sex-specific in the absence of contextual cues other than grammatical gender. This is especially strong for epicenes with masculine gender (i.e. *der Mensch* evokes almost only male associations). This indicates that grammatical gender can serve as a strong cognitive cue from which recipients deduce extralinguistic gender, even with the use of 'gender-neutral' nouns. As empirical research shows that the dissociation of grammatical and referential gender is no longer defensible, the question of how to address people who do not identify with the two 'traditional' genders becomes ever more pressing. Gender symbols, which are intended to encompass all gender identities (cf. Friedrich et al. 2021, Körner et al. 2022) and which a recent psycholinguistic study suggests to be the case (Zacharski & Ferstl 2023b), are a first attempt to solve this issue. They are, as we have described above, the target of many attacks on gender-inclusive language. The issue of 'gender-inclusive language', which actually comprises a large variety of strategies (pair forms, neutralisations, pronominal paraphrasings, gender symbols) is thus reduced to the issue of using gender symbols – and for many, this seems to be a 'yes' or 'no' question.[46] Two diametrically opposed positions seem irreconcilable – although we ought to assume there is a 'grey area' of people who do not actually care about the matter, but rather are, in a way, forced to state a definitive opinion (e.g. in polls). Similar to the linguistic 'battles' of the Bundestag outlined in this paper, the struggle about gender-inclusive forms goes beyond language, as it is closely tied to shifting social norms. It is about both self-denomination (as in the case of *schwul* and *lesbisch*) and (linguistically adequate) representations (as in the case of binomials and feminized forms). Sooner or later, this debate will also reach the Bundestag. If an MP uses a phonetic realization of gender symbols (e.g. a glottal stop between masculine base and feminine ending: *Politiker*innen -> Politiker[ʔ]innen*), then the stenographic transcription and later the protocol must accurately represent this. This is why discussions about the official inclusion of gender symbols in the transcription process arise (Hallik 2020: 86).[47]

---

[44] Cf. e.g. our talk on "Distribution of gender-inclusive orthographies in German press texts" at the "Linguistic Intersections of Language and Gender Conference" 2023; http://lilg.div-ling.org/wp-content/uploads/2023/06/Mueller-Spitzer_etal.pdf.

[45] Quoted from a joint statement of linguists (https://t1p.de/aedf) contesting the contribution „Wissenschaftsfremder Übergriff auf die deutsche Sprache" by Helmut Glück, in which he claims that ‚linguistics' is opposed to gender-inclusive language (Glück 2020).

---

[46] As the WDR poll 2023 (https://www1.wdr.de/nachrichten/gender-umfrage-infratest-dimap-100.html) shows, most respondents assess certain forms (especially binomials and collective nouns) positively. The only forms that receive more unfavourable than favourable opinions are gender symbols and their phonetic realization with a glottal stop. As Christine Olderdissen argued in her talk at the conference "Linguistic Intersection of Language and Gender" (Düsseldorf, 20-21 July 2023, http://lilg.div-ling.org/), the German press is in part responsible for this fixation on gender symbols and the general heatedness of the debate, as they have artificially kept the topic alive and nourished it with polemic, often unscientific contributions.

Another argument often discussed is about the legitimacy of 'language interventions', and the processes of 'natural' vs. 'artificial' language change (Haspelmath 2019). 'Interventions' are neither new nor 'dangerous' for language; rather, linguistic regulations in the parliament, as outlined above, have been widely accepted, and when they are not, it is usually political theatre. Language is nothing without its users, and competent speakers are in the position to adapt language to their needs – as is done in variable social, regional and cultural settings. Rarely has linguistic adaptation the potential to fuel such discussions as do gender symbols, which, if we took a closer look, is most likely connected to the underlying political and social implications of equality and diversity. In these debates, a consensus has already been reached, more or less tacitly (e.g. the use of binomials or gender-neutral terms instead of masculine generics), and the actual social issues that are tackled and highlighted by these linguistic means are lost out of sight (cf. Hark & Villa 2015). And instead of constructive discussions about linguistic means and the question of how queer identities can be made visible in language, the debate moves in circles, revolving around the 'ifs' instead of the 'hows'. As with earlier language debates, it is true that the "language struggle [...] to be observed is actually a cultural war"[48] (Simon 2022: 22). The issues under discussion are "really about the big picture, about the fundamental questioning of the foundations of our perception of the world" and "about emancipatory, but above all about post-essentialist experiences, interpretations and claims to participation."[49] (Simon 2022: 22)

---

[47] cf. also GfdS, the official linguistic counselling institution of the Bundestag: "If gender gaps [i.e. glottal stops] are used in spoken language as representations of gender asterisks or comparable forms, it is unclear how to properly transcribe them (e.g. in protocols of speeches)." (own translation; https://gfds.de/gendersternchen/)

## 5 Concluding remarks: The power of symbols

A recurring argument against gender-inclusive language holds that linguistic discussions serve as a sideshow distraction from issues which must be solved to achieve greater of gender equality. Aside from the fact that there is empirical evidence which suggests these areas are not necessarily separate issues (e.g. Jakiela & Ozier 2020), linguistic symbols have value in themselves. If we take the example of condolences after the death of a beloved person, we would not demand evidence that these condolences measurably changed the other person's feelings of grief in order to justify them. The point of this speech act is to acknowledge the other person's grief and to show compassion. This can also be true for non-discriminatory language. If a bus stop or street with the racist and colonialist term *Mohr* ('moor') in its name (e.g. *Mohrenstraße*) is renamed after years of discussion[50], this does not mean that PoCs are automatically less discriminated against in Germany. However, the renaming is a valuable symbolic act. Critical voices that highlight and discuss the colonialist and racist connotations of the term are heard and honoured in the public sphere with a sign visible to all. It is through such acts that discrimination can be tackled linguistically, and awareness of the underlying issues can be expressed. In the same way, by using gender-inclusive language with the symbols available, we can show that we care about gender equality and queer diversity. By tackling and changing linguistic forms like the generic masculine, we highlight real-world issues and raise awareness. By making these issues more salient, they become easier targets of change. Language plays such an important role in our individual and collective identity that it is hardly surprising that cultural upheavals such as the establishment of greater gender equality and diversity are also played out in the arena of language.

---

[48] Own translation, original: „Der zu beobachtende 'Sprachkampf' ist demnach eigentlich ein Kulturkampf."

[49] Own translation, original: „Denn es geht bei den verhandelten Fragen ' tatsächlich um das große Ganze, um die grundsätzliche Infragestellung von Grundlagen unserer Weltwahrnehmung' und ' um emanzipatorische, vor allem aber um post-essentialistische Erfahrungen, Deutungen und Partizipationsansprüche' "

It has further been argued that demands for more gender-inclusive language lead to restrictions of linguistic freedom and freedom of speech, and that is therefore not compatible with the concept of democracy (Eisenberg 2020: 15). However, neither the attempt of social minorities to gain more power and visibility, nor controversial debates in the public sphere is alien to democracy. Even if

> a constitution is formulated in a gender-fair way, this does not mean that there is language censorship or even language dictatorship. [...] The fact that all this has to be recorded at all is astonishing and shows the climate of political debate, in which any moderation seems to be lacking when it comes to gender-fair language[51] (Mangold 2021: 20).

---

[50] e.g. in Berlin: https://www.zdf.de/nachrichten/panorama/mohrenstrasse-berlin-umbenennung-100.html.

Similarly, the

> attempt to gain more influence is hardly *per se* illegitimate in democratic politics. And the claim that moralising minorities only want more power, while the majority is allegedly completely selfless in defending abstract ideals of universalism or individualism, is, to say the least, ideologically suspect. Those who do not want to talk about power usually have it themselves[52] (Müller 2021: 14).

The aim and moral obligation (cf. Stefanowitsch 2018: 41–42 for a discussion of the asymmetric distribution of possibilities to linguistically discriminate) of pluralist democracies should be to tolerate differences and to negotiate conflicts in a peaceful, respectful way.

As the language struggles of the past have made clear, political and parliamentary disputes about our world often correspond to disputes about words and labels. Who speaks about whom with which linguistic means, and who decides which forms of language are acceptable in which contexts – these are questions of power that have accompanied all linguistic-political disputes around shifting social norms.

---

[51] Own translation, original: Auch wenn „eine Verfassung geschlechtergerecht formuliert wird, so findet dadurch keine Sprachzensur oder gar Sprachdiktatur statt. […] Dass all dies überhaupt festgehalten werden muss, erstaunt und zeigt das Klima politischer Auseinandersetzung, in der teilweise jede Mäßigung zu fehlen scheint, wenn es um geschlechtergerechte Sprache geht."

[52] Own translation, original: Genauso ist „der Versuch, mehr Einfluss zu gewinnen, wohl kaum an sich illegitim in der demokratischen Politik. Und die Behauptung, die moralisierenden Minderheiten wollten nur mehr Macht, während man selber völlig uneigennützig mit der Verteidigung abstrakter Ideale von Universalismus oder Individualismus beschäftigt sei, ist, gelinde gesagt, ideologieverdächtig. Wer erst gar nicht über Macht reden will, hat sie meist selber inne."

Hirschauer, Stefan. 2020. Undoing Differences Revisited. Unterscheidungsnegation und Indifferenz in der Humandifferenzierung. *Zeitschrift für Soziologie* 49.5-6: 318–334. [https://doi.org/10.1515/zfsoz-2020-0027].

Hornscheidt, Antje. 2008. Die Konzeptualisierung von Gender in Komposita mit genderspezifizierenden Appellationsformen als erstem Glied. In Hornscheidt, Antje, *Gender Resignifiziert. Schwedische (Aus)Handlungen in und um Sprache*, pp. 207–270. Berlin: Nordeuropa-Institut (= Berliner Beiträge Zur Skandinavistik, Bd. 14).

Jakiela, Pamela & Owen W. Ozier. 2020. *Gendered Language* (IZA Discussion Paper No. 13126). [https://doi.org/10.2139/ssrn.3573296].

Jones, Lucy. 2021. Queer linguistics and identity: The past decade. *Journal of Language and Sexuality* 10.1: 13–24. [https://doi.org/10.1075/jls.00010.jon].

Kalverkämper, Hartwig. 1979. Die Frauen und die Sprache. *Linguistische Berichte* 62: 55–71.

Klein, Andreas. 2022. Wohin mit Epikoina? – Überlegungen zur Grammatik und Pragmatik geschlechtsindefiniter Personenbezeichnungen. In Gabriele Diewald, & Damaris Nübling (eds.), *Genus – Sexus – Gender*, pp. 135–190. Berlin, Boston: De Gruyter. [https://doi.org/10.1515/9783110746396-005].

Kolb, Matthias. 2013. Barack Obama: Kampfansage ans konservative Amerika. Süddeutsche.de. [https://www.sueddeutsche.de/politik/rede-des-us-praesidenten-zur-zweiten-amtszeit-obamas-kampfansage-ans-konservative-amerika-1.1579581] (accessed 22 January 2024).

Könne, Christian. 2018. Gleichberechtigte Mitmenschen? Deutschland Archiv from 07.09.2018. [https://www.bpb.de/themen/deutschlandarchiv/275113/gleichberechtigte-mitmenschen/] (accessed 22 January 2024).

Körner Anita, Bleen Abraham, Ralf Rummer & Fritz Strack. 2022. Gender Representations Elicited by the Gender Star Form. *Journal of Language and Social Psychology* 41.5: 1–19. [https://doi.org/10.1177/0261927X221080181].

Kotthoff, Helga & Damaris Nübling. 2018. *Genderlinguistik: Eine Einführung in Sprache, Gespräch und Geschlecht*. Tübingen: Narr Francke Attempto.

Krome, Sabine. 2020. Zwischen gesellschaftlichem Diskurs und Rechtschreibnormierung: Geschlechtergerechte Sprache als Herausforderung